\documentclass{article}
\usepackage{spconf,amsmath,amssymb,graphicx,booktabs}
\def\z{z^{-1}}

\title{Leaky-Integrator Reconstruction: Taming Error Accumulation\\
in Recursive Differenced Time-Series Forecasting}

\name{Zijiang Yang}
\address{New York University, Tandon School of Engineering, Brooklyn, NY 11201 \quad \texttt{zy3110@nyu.edu}}

\begin{document}
\ninept
\maketitle

\begin{abstract}
Recursive differenced forecasting, the standard remedy for non-stationarity, predicts one-step
changes and integrates them by cumulative summation. We show that this reconstruction is a
discrete integrator with a pole on the \emph{unit circle}, so the biased increment errors of a
learned nonlinear model are summed without bound and the rollout \emph{diverges}: at $336$
steps its normalised MAE reaches $1.6$--$3.8$ for every neural architecture tested, against
$0.80$ for a stable linear recursion. We then introduce \emph{leaky-integrator
reconstruction}, a training-free fix that moves the pole \emph{inside} the unit circle with
$H(z){=}1/(1{-}\gamma z^{-1})$, $\gamma{<}1$, bounding the accumulation of the model's own
increment errors. Applied post hoc with a \emph{single fixed} $\gamma{=}0.9$ (no retraining, a
two-line change to any deployed one-step or foundation-model forecaster), it beats the
traditional recursive integrator at every horizon, with the mean gain over seven diverging
architectures and twenty datasets \emph{growing} from $\sim\!3\%$ at $H{=}24$ to $23\%$ at
$H{=}96$, $37\%$ at $H{=}192$ and $51\%$ at $H{=}336$ (\textbf{$43$--$75\%$} across those
architectures; $78\%$ with an oracle pole), bringing all of them to $0.87$--$0.97$. Based on these extensive empirical experiments, adding a leaky integrator thus
improves recursive differenced time-series forecasting.
\end{abstract}


\section{Introduction}
\label{sec:intro}
A pervasive way to forecast a non-stationary series $\{y_t\}$ is to predict its first
difference $\Delta y_t=y_t-y_{t-1}$ and reconstruct levels by cumulative
summation~\cite{box1970time,engle1987co}. This underlies ARIMA, autoregressive and
probabilistic neural forecasters~\cite{salinas2020deepar}, recursively sampled foundation
models, and is kin to the re-centering of instance normalization~\cite{kim2022reversible}
and linear detrending~\cite{zeng2023transformers}. When such a forecaster is rolled out
\emph{recursively} (each predicted increment fed back to form the next input), the
reconstruction is a discrete integrator $1/(1-\z)$ with a pole at $z=1$, and the per-step
errors are integrated without bound. We show this is not benign: recursively rolled out
\emph{nonlinear} forecasters diverge, their $336$-step error growing several-fold above a
stable baseline.

We propose a direct remedy from classical signal processing: move the pole inside the unit
circle, i.e.\ reconstruct with a \emph{leaky} integrator $1/(1-\gamma\z)$, $\gamma<1$. The
fix is applied purely at reconstruction time (\textbf{no retraining, one scalar}), yet it
rescues every diverging architecture. Our contributions are:
(i)~a signal-processing account of recursive differenced forecasting as a unit-pole
integrator, and evidence that nonlinear rollout diverges (Sec.~\ref{sec:method},
\ref{sec:diag}); (ii)~the leaky-integrator reconstruction, which bounds the model's
error accumulation and, with a \emph{single fixed} pole and no tuning, cuts $H{=}336$ error by
$43$--$75\%$ across the diverging architectures over twenty datasets, the gain growing with
horizon (Sec.~\ref{sec:results}); and (iii)~a clear delineation of \emph{when} the remedy
applies (and, by design, does nothing), separating recursive over-accumulation from the
irreducible growth of stable and joint predictors (Sec.~\ref{sec:results}).

\section{Method: Leaky-Integrator Reconstruction}
\label{sec:method}
Let a one-step model predict the increment $\Delta\hat y$. Rolling it out and reconstructing
levels from the last observation $y_t$ gives
\begin{equation}
  \hat y_{t+h}=y_t+\sum_{j=1}^{h}\Delta\hat y_{t+j},
  \label{eq:recon}
\end{equation}
the discrete integrator $H(z)=1/(1-\z)$ with a pole at $z=1$. Writing the increment error
as $e_j=\Delta\hat y_{t+j}-\Delta y_{t+j}$, the reconstructed level error is the running sum
$\varepsilon_h=\sum_{j\le h}e_j$; for zero-mean, weakly correlated $e_j$ of variance
$\sigma^2$, $\operatorname{Var}(\varepsilon_h)\!\approx\!h\sigma^2$, so
$\mathrm{RMS}(\varepsilon_h)\!\propto\!\sqrt{h}$ for white errors; a learned model's errors are
\emph{biased and correlated}, so the sum accumulates faster. This excess is a property of the
integration, not of recursive feedback (Sec.~\ref{sec:diag}); the remedy is to damp the
integrator.

\noindent\textbf{Our reconstruction.} We replace the pure integrator with a \emph{leaky}
one, $H_\gamma(z)=1/(1-\gamma\z)$ with $0<\gamma\le1$:
\begin{equation}
  \hat y_{t+h}=y_t+s_h,\qquad s_h=\gamma\,s_{h-1}+\Delta\hat y_{t+h}.
  \label{eq:leaky}
\end{equation}
Older increments now decay geometrically as $\gamma^{h-j}$, moving the pole to $z=\gamma$
inside the unit circle. Equivalently, \eqref{eq:leaky} is a first-order IIR low-pass (an
exponentially-weighted integrator) on the predicted-increment stream, with $\gamma$
interpolating between pure integration ($\gamma{=}1$) and a one-increment forecast ($\gamma{=}0$).
The accumulation of the model's own errors is then \emph{bounded},
$\operatorname{Var}\!\to\!\sigma^2/(1-\gamma^2)$, but the damping also attenuates the true
increments: under a sustained drift $\mu$ the level bias grows as $\approx\!\mu h$ (a first-order
low-pass cannot track a ramp), so the pole helps when integrated model error dominates genuine
trend and not otherwise. Crucially \eqref{eq:leaky} is post-hoc: the increment model
is untouched, so the fix costs a single scalar and no retraining.

\begin{figure}[t]
  \centering
  \centerline{\includegraphics[width=\linewidth]{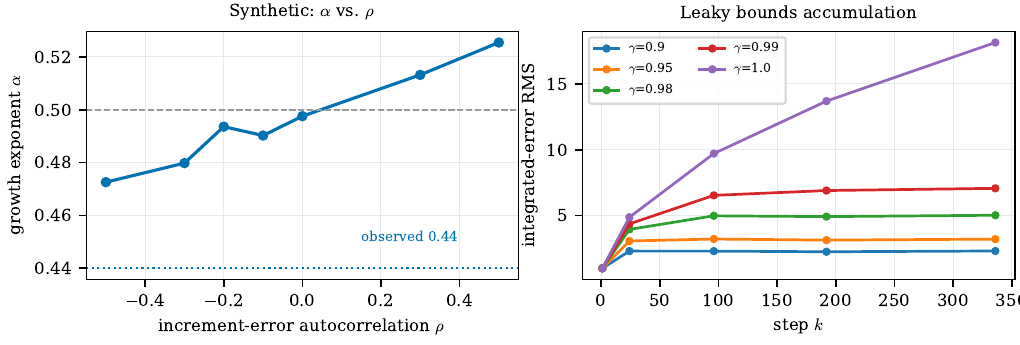}}
  \caption{Synthetic validation of the error model. \emph{Left:} integrating an AR(1)
  increment-error sequence of autocorrelation $\rho$ gives the fitted growth exponent $\alpha$; white
  noise yields $\alpha{=}0.5$, and short-memory correlation over $\rho\in[-0.5,0.5]$ moves it
  only within $0.47$--$0.53$. \emph{Right:} the leaky integrator caps the accumulated RMS at
  $1/\sqrt{1-\gamma^2}$, versus the unbounded $\sqrt{h}$ of the pure integrator.}
  \label{fig:syn}
\end{figure}

\noindent\textbf{Synthetic validation.} Fig.~\ref{fig:syn} verifies both halves of the
model directly. Integrating a white increment-error sequence gives $\alpha=0.50$; endowing the
errors with AR(1) autocorrelation $\rho\in[-0.5,0.5]$ shifts the fitted $\alpha$ only to
$0.47$--$0.53$, so short-memory correlation changes the constant, not the $\sqrt{k}$ law, and
the empirical $0.44$ (Sec.~\ref{sec:diag}) sits just below this band as a finite-horizon fit
effect. Passing the same sequence through the leaky filter caps its accumulated
RMS at $1/\sqrt{1-\gamma^2}$ ($5.0$ at $\gamma{=}0.98$, $2.3$ at $\gamma{=}0.9$), matching
theory to within $1\%$.

\section{Why Traditional Recursion Fails}
\label{sec:diag}
\begin{figure}[t]
  \centering
  \centerline{\includegraphics[width=\linewidth]{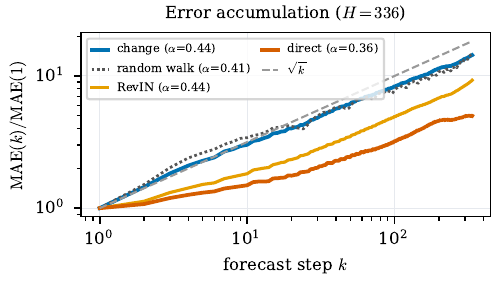}}
  \caption{Per-step error growth $\mathrm{MAE}(k)/\mathrm{MAE}(1)$ at $H{=}336$, averaged
  over six neural forecasters (LSTM, GRU, LSTM--GRU, DLinear, PatchTST, iTransformer), sixteen
  datasets and three seeds. Differenced (change) error tracks the $\sqrt{k}$ integrator law
  and coincides with the random-walk (irreducible) rate, as does RevIN-style re-centring;
  level (direct) prediction grows far more slowly.}
  \label{fig:growth}
\end{figure}

\begin{table}[t]
\caption{Per-step growth exponent $\alpha$ and factor $g(336)$ of differenced (change) vs.\
level (direct) forecasting, by model (benchmark, joint prediction). The change exponent is
$\approx\!\sqrt{k}$ for every architecture.}
\label{tab:alpha}
\centering
\begin{tabular}{lcccc}
\toprule
 & \multicolumn{2}{c}{change} & \multicolumn{2}{c}{direct} \\
\cmidrule(lr){2-3}\cmidrule(lr){4-5}
model & $\alpha$ & $g(336)$ & $\alpha$ & $g(336)$ \\
\midrule
LSTM         & 0.45 & 15.0 & 0.25 & 2.6 \\
GRU          & 0.45 & 14.8 & 0.26 & 3.2 \\
DLinear      & 0.41 & 13.6 & 0.49 & 13.0 \\
PatchTST     & 0.45 & 14.3 & 0.30 & 3.4 \\
iTransformer & 0.43 & 14.1 & 0.27 & 2.7 \\
\bottomrule
\end{tabular}
\end{table}

The integrator makes error grow as $\sqrt{k}$ with forecast step $k$. On a benchmark of five
forecasters over sixteen datasets and seven horizons, the change formulation grows with
exponent $\alpha{=}0.44$ (Fig.~\ref{fig:growth}), reaching $14\times$ at $336$ steps versus
$5.0\times$ for level prediction; the exponent is architecture-independent
($\alpha\in[0.41,0.45]$, Table~\ref{tab:alpha}) and coincides with the random-walk baseline ($\alpha{=}0.41$), the
\emph{irreducible} rate of a random walk. This $\sqrt{k}$ is the floor for
\emph{white} errors; a learned model's increment errors are biased and correlated, so
integrating them sends $H{=}336$ normalised MAE to $1.6$--$3.8$ (pure column, Table~\ref{tab:main}),
well above the $\approx\!0.8$ floor. Crucially this excess is intrinsic to the \emph{integration},
not to feedback: a \emph{teacher-forced} rollout (every increment predicted from the true past,
no feedback) accumulates \emph{as much or more} (mean $H{=}336$ nMAE over MLP/GRU/Transformer:
$2.3$ teacher-forced vs.\ $1.8$ recursive), and free-running recursion even self-damps as its
inputs contract. Damping the integrator removes this excess, leaving the irreducible $\sqrt{k}$.

\section{Experimental Setup}
\label{sec:setup}
\noindent\textbf{Data.} We use twenty univariate target series spanning ten domains and a
wide range of sampling rates, from daily financial data to $10$--$15$\,min energy, climate
and traffic sensors: cryptocurrencies (BTC, ETH, SOL, BNB, XRP), equity indices
(S\&P\,500, Nikkei), commodities (oil, gold), rates/volatility (10y yield, VIX), FX
(exchange rate), energy (electricity), climate (weather, temperature), traffic, and
industrial ETT sensors (ETTh1/2, ETTm1/2). Each series is split $70/10/20$
\emph{chronologically} (train/val/test) to preclude look-ahead leakage, and
$z$-standardised using only training-window statistics so that normalised MAE is scale-free
and comparable across domains. We group the series into three regimes: \emph{drift} (12
trending financial/commodity series, where differencing is needed), \emph{structured} (4 ETT
sensors with strong daily/weekly seasonality), and \emph{near-stationary} (4: electricity,
weather, temperature, traffic), which isolate where the integrated errors drift most.

\noindent\textbf{Architectures.} We evaluate nine one-step increment predictors: linear
autoregression, MLP, LSTM~\cite{hochreiter1997long}, GRU, DLinear~\cite{zeng2023transformers},
a dilated TCN, PatchTST~\cite{nie2023patchtst}, iTransformer~\cite{liu2024itransformer} and a
small Transformer. Our LSTM, GRU and Transformer rollouts are genuine autoregressive decoders
(the native mode of ARIMA, DeepAR and sampled foundation models), where exposure
bias~\cite{bengio2015scheduled} is expected; DLinear, PatchTST and iTransformer are natively
\emph{joint} multi-output models that we additionally roll out to test whether the mechanism
is architecture-general (their joint use is inert, Sec.~\ref{sec:results}). These are compact
univariate reimplementations that keep each design's core (patching, inverted attention,
linear decomposition); the univariate iTransformer in particular loses its cross-variate
mixing, so its rollout figures probe the mechanism rather than rank the architecture.

\noindent\textbf{Training.} Every model is trained once (single seed) to predict the next
normalised increment $\Delta z_t$ from the previous $L{=}96$ increments, minimising a one-step
MSE with Adam (learning rate $10^{-3}$, batch $256$, a few epochs over the sliding windows of
the training split, sub-sampled for the longest series). The networks are deliberately
compact, so that the reconstruction pole, not model capacity, is the variable under
study, with a $64$-unit MLP, $32$--$64$-unit recurrent cells, a three-block dilated TCN, a
one-layer width-$32$ four-head Transformer, and lightweight univariate PatchTST/iTransformer
encoders. The direct multi-horizon (direct-MH) variant of each shares this backbone but
replaces the one-step head with an $H$-output head trained jointly on all $H$ future
increments.

\noindent\textbf{Rollout, reconstruction and metric.} At each of up to $200$ stride-spaced
forecast origins per test series (series yielding fewer than $15$ valid origins are dropped),
we roll the one-step model out autoregressively to $H{=}336$, feeding each predicted
increment back as the next input, and reconstruct levels from the last observation either
with the traditional pure integrator ($\gamma{=}1$, a cumulative sum) or with our leaky filter
(Eq.~\eqref{eq:leaky}, a first-order IIR). We report \emph{normalised MAE}: the absolute level
error in units of the training standard deviation, $|\hat y_{t+k}-y_{t+k}|/\sigma_{\text{train}}$
(scale-free and averageable across series), averaged over steps $k{=}1,\dots,H$ for each
horizon $H\in\{24,96,192,336\}$, then over origins, then over datasets. For the oracle upper bound the pole
is swept over a $13$-point grid in $[0,1]$; unless stated we use the deployable $\gamma{=}0.9$.

\section{Results}
\label{sec:results}
\begin{table}[t]
\caption{Per-dataset normalised MAE at $H{=}336$ (mean over the seven nonlinear
architectures). Our fixed-$\gamma{=}0.9$ reconstruction lowers error over the traditional
recursive integrator (rec.) on \emph{nineteen of twenty} datasets, dramatically where
recursion diverges (e.g.\ Traffic $7.07\!\to\!1.06$, Weather $2.93\!\to\!1.06$), the sole
exception being the near-random-walk SP500 ($0.90\!\to\!0.96$). \textbf{Bold} marks the better
of rec.\ and ours; a well-trained direct multi-horizon model (d-MH) is shown for reference.}
\label{tab:perds}
\centering
\begingroup\setlength{\tabcolsep}{4pt}
\begin{tabular}{lccc@{\hskip 1.6em}lccc}
\toprule
dataset & rec. & ours & d-MH & dataset & rec. & ours & d-MH \\
\midrule
BTC      & 2.81 & \textbf{1.33} & 1.37 & VIX      & 1.27 & \textbf{0.59} & 0.60 \\
ETH      & 2.05 & \textbf{0.71} & 0.74 & Exch.    & 0.68 & \textbf{0.47} & 0.48 \\
SOL      & 1.98 & \textbf{0.67} & 1.02 & ETTh1    & 1.16 & \textbf{0.34} & 0.34 \\
BNB      & 1.97 & \textbf{0.76} & 0.67 & ETTh2    & 1.90 & \textbf{0.63} & 0.54 \\
XRP      & 4.09 & \textbf{3.43} & 3.51 & ETTm1    & 1.01 & \textbf{0.28} & 0.27 \\
SP500    & \textbf{0.90} & 0.96 & 0.92 & ETTm2    & 2.07 & \textbf{0.56} & 0.46 \\
Nikkei   & 0.84 & \textbf{0.81} & 0.79 & Elec.    & 1.42 & \textbf{1.06} & 0.76 \\
Oil      & 0.95 & \textbf{0.49} & 0.50 & Weather  & 2.93 & \textbf{1.06} & 1.01 \\
Gold     & 0.99 & \textbf{0.94} & 0.88 & Temp     & 2.71 & \textbf{1.22} & 1.23 \\
Yield10y & 1.05 & \textbf{0.57} & 0.65 & Traffic  & 7.07 & \textbf{1.06} & 1.16 \\
\bottomrule
\end{tabular}
\endgroup
\end{table}
\begin{table}[t]
\caption{\textbf{Our leaky reconstruction vs.\ the traditional pure-recursive integrator}, at
$H{=}336$ (normalised MAE, $20$ datasets, nine architectures). ``ours'' is a \emph{single
fixed} pole $\gamma{=}0.9$ (deployable, no tuning; $\gamma{=}1$ for stable linear-AR);
``oracle'' is the best per-dataset pole (an upper bound); ``Impr.'' is the fixed-$\gamma$
reduction over the pure integrator. Nonlinear recursion diverges and the fixed pole rescues
it; stable linear-AR is correctly unchanged.}
\label{tab:main}
\centering
\begingroup\setlength{\tabcolsep}{6pt}
\begin{tabular}{lcccc}
\toprule
architecture & recursive & \textbf{ours} & oracle & Impr. \\
\midrule
linear-AR    & 0.80 & 0.80 & 0.79 & \phantom{0}0\% \\
MLP          & 1.56 & \textbf{0.90} & 0.84 & 43\% \\
LSTM         & 1.57 & \textbf{0.90} & 0.86 & 43\% \\
GRU          & 1.77 & \textbf{0.88} & 0.82 & 50\% \\
DLinear      & 0.90 & \textbf{0.86} & 0.81 & \phantom{0}4\% \\
TCN          & 1.75 & \textbf{0.88} & 0.86 & 50\% \\
PatchTST     & 1.74 & \textbf{0.87} & 0.82 & 50\% \\
iTransformer & 1.74 & \textbf{0.89} & 0.81 & 49\% \\
Transformer  & 3.80 & \textbf{0.97} & 0.84 & 75\% \\
\bottomrule
\end{tabular}
\endgroup
\end{table}

\begin{figure}[t]
  \centering
  \includegraphics[width=\linewidth]{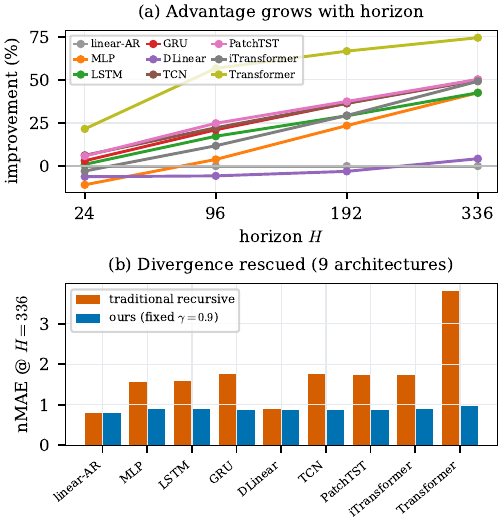}
  \caption{\emph{(a)} Improvement of our leaky reconstruction (fixed $\gamma{=}0.9$) over the
  traditional pure-recursive integrator vs.\ horizon, the advantage grows monotonically with
  $H$ for every nonlinear architecture and is flat for stable linear-AR. \emph{(b)} At
  $H{=}336$ the pure integrator diverges (nMAE $1.6$--$3.8$); our method restores every
  diverging architecture to $\approx\!0.87$--$0.97$.}
  \label{fig:rescue}
\end{figure}

\begin{figure}[t]
  \centering
  \centerline{\includegraphics[width=\linewidth]{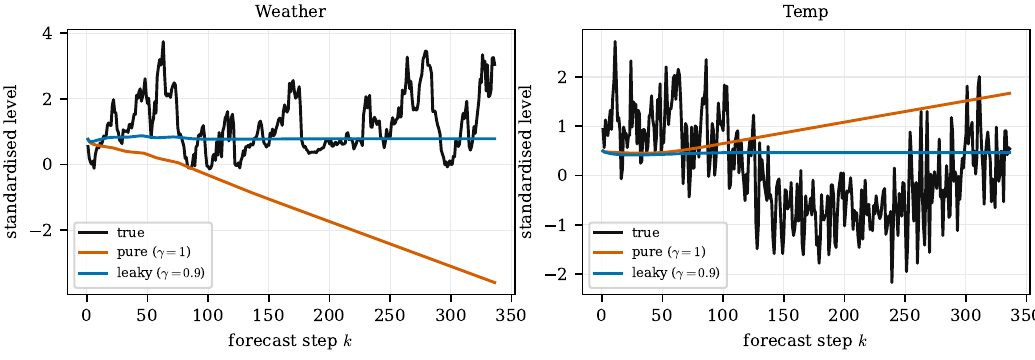}}
  \caption{A single $336$-step recursive Transformer forecast on two near-stationary
  series. The pure integrator ($\gamma{=}1$) drifts away, down on Weather, up on
  Temp, as the integrated increment errors accumulate, while the leaky reconstruction
  ($\gamma{=}0.9$) stays anchored to the true level.}
  \label{fig:traj}
\end{figure}

\noindent\textbf{Improvement over traditional recursion (nine architectures).} We train a
one-step increment predictor for each of nine architectures (linear-AR, MLP, LSTM, GRU,
DLinear, TCN, PatchTST, iTransformer, Transformer), roll each out to $H{=}336$ on twenty
datasets, and reconstruct with the pure ($\gamma{=}1$) versus our leaky pole.
Table~\ref{tab:main} and Fig.~\ref{fig:rescue} give the result. Every nonlinear model
diverges under the pure integrator; a \emph{single fixed} pole $\gamma{=}0.9$ pulls all of
them back to $\approx\!0.87$--$0.97$ nMAE, a \textbf{$43$--$75\%$ error reduction} (the
Transformer, which diverges worst at $3.8$, gains most; an oracle per-dataset pole reaches
$0.81$--$0.86$, i.e.\ up to $78\%$); Table~\ref{tab:perds} gives the full per-dataset
breakdown. Linear-AR is stable and, correctly, left at $\gamma{=}1$.
Fig.~\ref{fig:traj} shows the mechanism on individual forecasts, the pure integrator
drifts away while our reconstruction stays anchored, and Table~\ref{tab:regime} breaks the
effect down by regime: divergence is worst on \emph{near-stationary} series (pure nMAE
$3.53$, where a trendless target gives the integrated increment errors no real trend to anchor to), which the fixed
pole rescues to $1.10$. Over three seeds the fixed-$\gamma{=}0.9$ error is stable (std $\le
0.24$ nMAE) while the pure integrator is seed-erratic (Transformer std $5.4$), so the fix also
removes recursion's run-to-run instability.

\begin{table}[t]
\caption{Recursive rollout by regime at $H{=}336$ (nMAE, mean over the seven nonlinear
architectures). The pure integrator diverges most on near-stationary series; our
reconstruction rescues every regime, approaching a well-trained direct multi-horizon model
(direct-MH, shown for reference) at no training cost.}
\label{tab:regime}
\centering
\begin{tabular}{lccc}
\toprule
regime & recursive & \textbf{ours} & direct-MH \\
\midrule
drift ($n{=}12$)     & 1.63 & \textbf{0.98} & 1.01 \\
structured ($n{=}4$) & 1.54 & \textbf{0.45} & 0.40 \\
stationary ($n{=}4$) & 3.53 & \textbf{1.10} & 1.04 \\
\bottomrule
\end{tabular}
\end{table}

\begin{table}[t]
\caption{Improvement of our reconstruction (fixed $\gamma{=}0.9$) over the traditional
recursive integrator (\% reduction in nMAE) \emph{by horizon}. The advantage grows
monotonically with $H$; at short horizons the fixed pole can slightly over-damp (negative
entries), as $\gamma^\ast\!\to\!1$ when $H{\to}0$ (Fig.~\ref{fig:gamma}). Linear-AR is held
at $\gamma{=}1$.}
\label{tab:horizon}
\centering
\begin{tabular}{lcccc}
\toprule
architecture & $H{=}24$ & $H{=}96$ & $H{=}192$ & $H{=}336$ \\
\midrule
linear-AR    &   0 &  0 &  0 &  0 \\
MLP          & $-11$ &  4 & 23 & 43 \\
LSTM         &   1 & 17 & 29 & 43 \\
GRU          &   3 & 21 & 36 & 50 \\
DLinear      & $-6$ & $-6$ & $-3$ &  4 \\
TCN          &   6 & 22 & 36 & 50 \\
PatchTST     &   6 & 25 & 38 & 50 \\
iTransformer & $-3$ & 12 & 29 & 49 \\
Transformer  &  22 & 57 & 67 & 75 \\
\bottomrule
\end{tabular}
\end{table}

\noindent\textbf{The advantage grows with the horizon.} Fig.~\ref{fig:rescue}(a) and
Table~\ref{tab:horizon} show the improvement is negligible at short horizons (where little has
accumulated, and a fixed pole can even over-damp) and rises monotonically to $43$--$75\%$ at
$H{=}336$ for the recurrent and attention models, so the method is most valuable exactly at the
long horizons where naive recursion is most damaged.

\begin{figure}[t]
  \centering
  \centerline{\includegraphics[width=\linewidth]{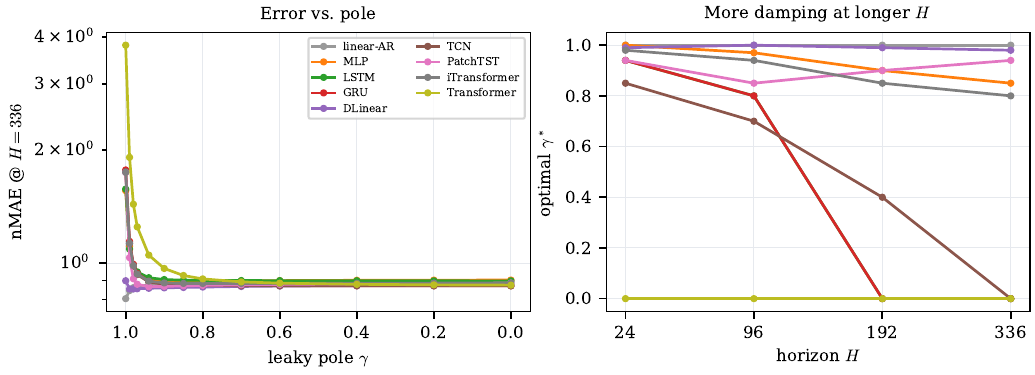}}
  \caption{Setting the pole. \emph{Left:} $H{=}336$ error vs.\ $\gamma$ ($\gamma$ decreasing
  rightward, log scale), nonlinear models fall steeply and plateau once
  $\gamma\!\lesssim\!0.9$; stable linear-AR degrades under any damping. \emph{Right:} the
  optimal pole $\gamma^\ast$ falls with the horizon for most architectures.}
  \label{fig:gamma}
\end{figure}

\noindent\textbf{The pole needs no tuning.} Fig.~\ref{fig:gamma} shows the error is a smooth
function of $\gamma$ that plateaus once $\gamma\!\lesssim\!0.9$, with the optimum shifting to
stronger damping at longer horizons. This is why we headline a \emph{single fixed}
$\gamma{=}0.9$: it recovers almost all of the oracle gain ($+43\%$ vs.\ $+46\%$ for MLP;
$+75\%$ vs.\ $+78\%$ for the Transformer, Table~\ref{tab:main}) with no per-dataset
search. In a separate run using the earlier half of the post-training origins for validation
and the later half for test, selecting $\gamma$ per (model,dataset) reaches test nMAE $0.76$,
just below the fixed $\gamma{=}0.9$ ($0.81$ on those origins) and near the
oracle ($0.73$): the fixed default leaves little on the table. The one rule: withhold damping from a \emph{stable} recursion, where $\gamma{=}0.9$ instead
\emph{costs} accuracy (linear-AR: $+9.3\%$), a case the super-$\sqrt{k}$ diagnostic
(Sec.~\ref{sec:disc}) flags.

\noindent\textbf{Relation to direct multi-horizon prediction.} A well-trained \emph{direct}
multi-horizon (direct-MH) head, emitting all $H$ steps jointly, is overall competitive with our
reconstructed one-step model (Tables~\ref{tab:perds},~\ref{tab:regime}) but needs its own
model and training run; ours reuses an existing predictor, so the two are complementary.

\noindent\textbf{Scope.} The remedy targets integration \emph{over}-accumulation, not the
irreducible $\sqrt{k}$: it is inert for stable linear recursion and for \emph{joint}
forecasters, which never integrate increments ($\approx\!0\%$ gain).

\section{Discussion}
\label{sec:disc}
\noindent\textbf{Practical guidance and cost.} Deployment is a two-line change (an $O(H)$
first-order filter over the predicted increments, fixed $\gamma{=}0.9$, no parameters or
retraining), gated by the super-$\sqrt{k}$ diagnostic on a validation split so that stable or
joint models are left at $\gamma{=}1$.

\section{Relation to Prior Work}
\label{sec:prior}
\noindent\textbf{Differencing and stationarisation.} Differencing to induce stationarity is
the foundation of Box--Jenkins ARIMA modelling~\cite{box1970time}, and error-correction /
cointegration models~\cite{engle1987co} formalise how integrated series are recombined. The
same idea reappears in deep forecasting as per-window re-centring: reversible instance
normalisation~\cite{kim2022reversible}, non-stationary
attention~\cite{liu2022nonstationary} and DLinear's linear detrending~\cite{zeng2023transformers}
all forecast a de-trended or differenced target, exactly the setting we analyse; the leaky
reconstruction we adopt is itself a first-order exponential smoother of the increment
stream~\cite{gardner2006exponential}, a post-hoc relative of Gardner and McKenzie's damped-trend
method~\cite{gardner1985forecasting}, which was designed to damp uncertain long-range trend
extrapolation.

\noindent\textbf{Long-horizon architectures and the multi-horizon strategy.} Most modern
long-sequence forecasters (Informer~\cite{zhou2021informer},
Autoformer~\cite{wu2021autoformer}, FEDformer~\cite{zhou2022fedformer},
PatchTST~\cite{nie2023patchtst} and iTransformer~\cite{liu2024itransformer}) emit all $H$
steps \emph{jointly} (direct multi-horizon), and N-BEATS~\cite{oreshkin2020nbeats} and its
hierarchical successor N-HiTS~\cite{challu2023nhits} use direct basis-expansion heads,
precisely to \emph{avoid} recursion. Whether to forecast directly or iterate one step at a
time is a long-standing question in the forecasting literature~\cite{marcellino2006comparison}.
Recursive rollout nonetheless remains the native mode for classical ARIMA, autoregressive
probabilistic models such as DeepAR~\cite{salinas2020deepar}, and autoregressively sampled
foundation models such as TimesFM~\cite{das2024timesfm}, the regime in which our fix is most
needed.

\noindent\textbf{Exposure bias and our position.} In recurrent sequence
models~\cite{sutskever2014sequence} recursive error growth is known as exposure bias and is
usually attacked at \emph{training} time (scheduled sampling~\cite{bengio2015scheduled},
professor forcing~\cite{lamb2016professor} and probabilistic
rollout~\cite{salinas2020deepar}), while its bias--variance trade-off has been studied for
multistep forecasting~\cite{taieb2016bias}. In contrast, we identify recursive differenced
reconstruction as a \emph{unit-pole integrator} and fix it by \emph{training-free} pole
placement; to our knowledge this reconstruction and its error analysis are new to time-series
forecasting.

\section{Conclusion}
\label{sec:concl}
This paper makes three contributions. \emph{(i)~Diagnosis.} Differenced forecasting
reconstructs levels with a unit-pole integrator, so a learned model's biased, correlated
increment errors are integrated without bound; on twenty datasets every recursively rolled-out
nonlinear architecture diverges (up to $3.8$ normalised MAE at $H{=}336$), and teacher forcing
shows the cause is the integration itself, not recursive feedback. \emph{(ii)~Remedy.} Placing the pole \emph{inside} the unit
circle, a leaky integrator with one scalar $\gamma$, bounds the model's own error accumulation
at $\sigma^2/(1-\gamma^2)$; applied post hoc with a fixed $\gamma{=}0.9$ and no retraining, it
cuts $336$-step error by $43$--$75\%$ on the seven diverging architectures, the gain growing
with the horizon, a tuned $\gamma$ adding little ($0.76$ vs.\ $0.81$) and seed variability
falling from $5.4$ to $0.24$. \emph{(iii)~Scope.} The analysis also says when to
withhold the fix: it is inert for joint predictors, which never integrate increments, and
hurts stable recursions (linear-AR $+9.3\%$), a case the super-$\sqrt{k}$ diagnostic flags. The
remedy is a two-line change to any deployed one-step or foundation-model forecaster.

\bibliographystyle{IEEEbib}
\bibliography{refs}

\end{document}